\documentclass[lettersize,journal]{IEEEtran}
\usepackage{amsmath,amsfonts}
\usepackage{algorithmic}
\usepackage{algorithm}
\usepackage{array}
\usepackage[caption=false,font=normalsize,labelfont=sf,textfont=sf]{subfig}
\usepackage{textcomp}
\usepackage{stfloats}
\usepackage{url}
\usepackage{verbatim}
\usepackage{graphicx}
\usepackage{cite}
\begin{document}

\title{Flow-Guided Controllable Line Drawing Generation}

\author{Chengyu Fang, Xianfeng Han\IEEEauthorrefmark{1},~\IEEEmembership{Member,~IEEE,}
\thanks{Cheng-Yu Fang, Xian-Feng Han with the College of Computer and Information Science, Southwest University, Beibei District, Chongqing, China e-mail: cnyvfang@email.swu.edu.cn, xianfenghan@swu.edu.cn.}
\thanks{\IEEEauthorrefmark{1} Corresponding Author.}
\thanks{Manuscript received April 19, 2023; revised August 16, 2023.}}

\markboth{SUBMITTED TO IEEE TRANSACTIONS ON CIRCUITS AND SYSTEMS FOR VIDEO TECHNOLOGY}%
{Shell \MakeLowercase{\textit{et al.}}: A Sample Article Using IEEEtran.cls for IEEE Journals}


\maketitle

\begin{abstract}
In this paper, we investigate the problem of automatically controllable artistic character line drawing generation from photographs by proposing a Vector Flow Aware and Line Controllable Image-to-Image Translation architecture, which can be viewed as an appealing intersection between Artificial Intelligence and Arts. Specifically, we first present an Image-to-Flow network (I2FNet) to efficiently and robustly create the vector flow field in a learning-based manner, which can provide a direction guide for drawing lines. Then, we introduce our well-designed Double Flow Generator (DFG) framework to fuse features from learned vector flow and input image flow, guaranteeing the spatial coherence of lines. Meanwhile, to allow for controllable character line drawing generation, we integrate a Line Control Matrix (LCM) into DFG and train a Line Control Regressor (LCR) to synthesize drawings with different styles by elaborately controlling the level of details, such as thickness, smoothness, and continuity, of lines. Finally, we design a Fourier Transformation Loss to further constrain the character line generation from the frequency domain view of the point. Quantitative and qualitative experiments demonstrate that our approach can obtain superior performance in producing high-resolution character line-drawing images with perceptually realistic characteristics. 
\end{abstract}

\begin{IEEEkeywords}
Image-to-Image Translation, Controllable Generation, Character Line Drawing, Edge Tangent Flow
\end{IEEEkeywords}

\section{Introduction}
\label{sec:intro}
\IEEEPARstart{C}{haracter} line drawing or line art refers to a process that uses various kinds of lines (e.g., straight lines or curved lines) to create an abstract and stylistic illustration for the visual properties of characters, including cartoons, manga and real people. It can be actually considered as a concise yet effective art modality in the domain of non-photorealistic rendering (NPR) \cite{wang2012abstract} communicating geometric shape and semantic information to viewers \cite{chan2022learning}. Therefore, character line drawing spans a broad spectrum of applications, such as colourization \cite{kim2022late}, 2D animation \cite{mo2021general} and image editing \cite{alaluf2022hyperstyle}. However, free-hand character line drawing requires professional drawing skills, expensive labour costs and higher time consumption \cite{yi2020unpaired}. It is highly desirable to develop automatic techniques, aiming to generate character line drawings from given photographs automatically.

Early traditional approaches for performing photo-to-line drawing translation mainly depend on low-level edge-based techniques (e.g., Canny edge detector), which focus on using the gradient to capture accurate edges with less illustration of artistic stylization \cite{kang2007coherent}. Yet what we want is a perceptually meaningful image. Meanwhile, for the persons in photos, not only the outlines but also the implicit properties like the material of the clothes \cite{kang2007coherent} should be taken into consideration to form high-quality illustrations. Fortunately, in recent years, due to continuing increase in the learning power of hierarchical representation, deep learning techniques, especially the emergence of generative adversarial networks (GANs) \cite{goodfellow2020generative} have revolutionized the area of image-to-image (I2I) translation. Many works have explored the production of line arts using deep convolutional neural networks (CNNs), but they mainly centre on drawings for face portraits \cite{yi2020line} and objects \cite{bhunia2021vectorization}. 

On the other hand, most previous methods of I2I translation tasks actually are not ideal choices for high-quality character line drawing generation. This is mainly due to the following challenges: (1) line drawing is a fairly abstract representation, which uses a set of much sparser lines to depict the visual characteristics of characters, especially the body part.  (2) Lacking of direction guidance, the lines resulting from these models are usually unsharp, incomplete, unnatural and noisy. (3) Few methods are able to provide line detail adjustment in a user-controllable way. 

To address the above problems, in this work, we propose a vector flow aware and line controllable network architecture based on the core idea of GAN to generate even higher visual-quality character line drawings. First, we use the learning-based model to produce vector flow with better preservation of edge direction efficiently. Then, under the guidance of the learned flow, together with the original input image flow, our Double Flow Generator (DFG) can improve the coherence of lines and reduce noise by fusing features from both flows. Finally, control over different levels of line details is achieved by the introduction of our Line Control Matrix and Line Control Regressor. With these well-designed modules, strategies and supervision by GAN loss, Line Control loss, Pixel-wise loss as well as our developed Fourier Transformation loss, our method can capture coherent, stylistic and clean lines from photographs of the persons.

In summary, our main contributions are as follows:
\begin{itemize}
    \item An Image-to-Flow network (I2FNet) is designed for efficient generation of the edge flow field, which can maintain the direction of the edge lines.
    
    \item We devise a Double Flow Generator (DFG) model to enhance the spatial coherence of character lines by mutually fusing information flow from the learned vector field above and the original input image.
    
    \item We obtain control over different levels of line detail by embedding a Control Matrix into DFG and training a Control Regressor to produce different character line drawing styles. 
    
    \item We construct a GAN-based Vector Flow Aware Image-to-Image framework for automated generation of visually high-quality character line drawings from photographs, and introduce a Fourier Transformation Loss to supervise the learning process in terms of frequency domain.  
    
    \item Both quantitative and qualitative comparisons against state-of-the-art methods show the effectiveness, competitiveness, or even superiority of our proposed architecture.
\end{itemize}

\section{Related Work}
Character line drawing/art representation visualizes the appearance of a person/cartoon/manga in images or photographs using a set of lines. Its automated generation can be considered one of the most important applications of Image-to-Image (I2I) translation in arts. Here, we briefly review related work in these fields.  
 
\subsection{Image-to-Image Translation}
The recent increase in the image generation power of  Generative Adversarial Networks (GANs) has created unprecedented opportunities for the development of the Image-to-Image (I2I) translation field, whose  objective is to learn a mapping from the source image domain to the target image domain. \cite{he2023weaklysupervised} \cite{he2023HQG} \cite{he2019image} A broad spectrum of successful applications of I2I translation can be found in pencil drawing \cite{li2019im2pencil} and image colourization \cite{lee2020reference} \cite{wang2022palgan}.

Pix2Pix \cite{isola2017image} adopts the conditional GANs to formulate the first general-purpose image-to-image translation structure suitable for graphics and vision tasks. However, it has difficulty producing high-resolution and realistic images. To address this problem, Wang et al. \cite{wang2018high} used a coarse-to-fine generator, a multi-scale discriminator and an improved adversarial loss to form the high-resolution version of Pix2Pix, named Pix2PixHD. CycleGAN \cite{zhu2017unpaired} introduces two adversarial discriminators and two cycle-consistency losses to achieve mappings from $X$ domain to $Y$ domain and from $Y$ to $X$. While Zhu et al. \cite{zhu2017toward} proposed the BiCycleGAN to create a bijective connection between latent space and  output by combining both the conditional variational autoencoder GAN (cVAE-GAN) and conditional latent regressor GAN (cLR-GAN), and it can obtain more realistic and diverse results. In order to avoid washing away information, a spatially-adaptive denormalization layer \cite{park2019semantic} \cite{he2023strategic}  is designed. With its help, the semantic information can be propagated through the network via the learned transformation for photorealistic image generation. 

\subsection{Line Drawing Generation}
Line drawing can be viewed as a special kind of art modality, which attempts to represent the visual properties of an object or character or scene using basic lines without shading. However, generating line drawing is a challenging study due to its characteristics of abstraction, diversity, sparsity and invariance \cite{xu2022deep} \cite{He2023Camouflaged} .

Li et al. \cite{li2017deep} developed an encoder-decoder network to achieve structural line extraction from screen-rich manga images. Xiang et al. \cite{xiang2022adversarial} proposed a framework containing two generators and two discriminators that perform photo-to-sketch and sketch-to-photo learning jointly. For artistic portrait drawing generation, APDrawingGAN \cite{yi2019apdrawinggan} makes use of a global network for global facial structure description, and six local networks for eyes, nose, mouth, hair and background generation. However, it only produces coarse drawings. To enhance fine details, APDrawingGAN++ \cite{yi2020line} chooses to exploit an autoencoder as a generator, and integrates a hair classier, a lip classifier to avoid synthesizing undesirable style. Yi et al. \cite{yi2022quality} devised an asymmetric cycle GAN framework whose learning process is guided by a novelly designed quality metric for better-looking APDrawings generation. Chan et al. \cite{chan2022learning} decoupled line drawings into encoding geometry, semantics and appearance enforced by CLIP, appearance and geometry losses. 

Different from these methods, we mainly pay attention to controllable artistic character line drawing synthesis from full body images/photographs, involving cartoons and real persons, which contains a much sparser set of lines, especially in the body part. Our study leverages feature representations from the image and edge flows to synthesize coherent, stylistic and clear lines under the supervision of Adversarial loss, pixel-wise loss, and FFT loss. The introduction of the Line Control Matrix and Line Control Regressor obtains control over line details.

\section{Data Preparation}
\label{datapre}
In this study, we formulate both Image-to-Flow and Image/Photograph-to-Character line drawing transformations as supervised Image-to-Image translation tasks. Therefore, in order to find the mapping from character image/photograph to edge tangent flow field and that from image to character line drawings in a learning-based manner, building a high-quality dataset is desirable for training and evaluating our model.

We collect a large number of character images from online websites, which mainly contain images/photographs of men, women, manga/cartoon boys and girls to ensure the diversity and richness of data. In order to make these images more suitable for processing by deep learning networks, we first scale them to a uniform size of $1024 \times 1024$ pixels.       

Then, for each image, we use the traditional edge tangent flow acquisition strategy \cite{kang2007coherent} to generate its corresponding ETF vector field. On the other hand, since it is fairly difficult to obtain character image/photograph-line drawing pairs, we choose to use Flow-based Difference of Gaussian methods to produce several smoothing line drawings with different levels of details for every image, and record the corresponding line control parameter. Finally, we group the images and their corresponding ETF files as well as several line drawings with different control parameters to construct the final dataset. Finally, our dataset contains 1,634 images in total, where 1,037 samples are selected for training, and 327 images for testing. In the case of controllable generation, five line drawings are selected for each image, together with the corresponding ETF to train our character image/photograph-to-line drawing translation network. In addition, it should be noted that in order to make a fair comparison, in our comparison experiments and ablation study, we use only one character line drawing for every image to train all the models, which is generated using a fixed control parameter value.

\begin{figure*}
\centering
\includegraphics[width=\textwidth]{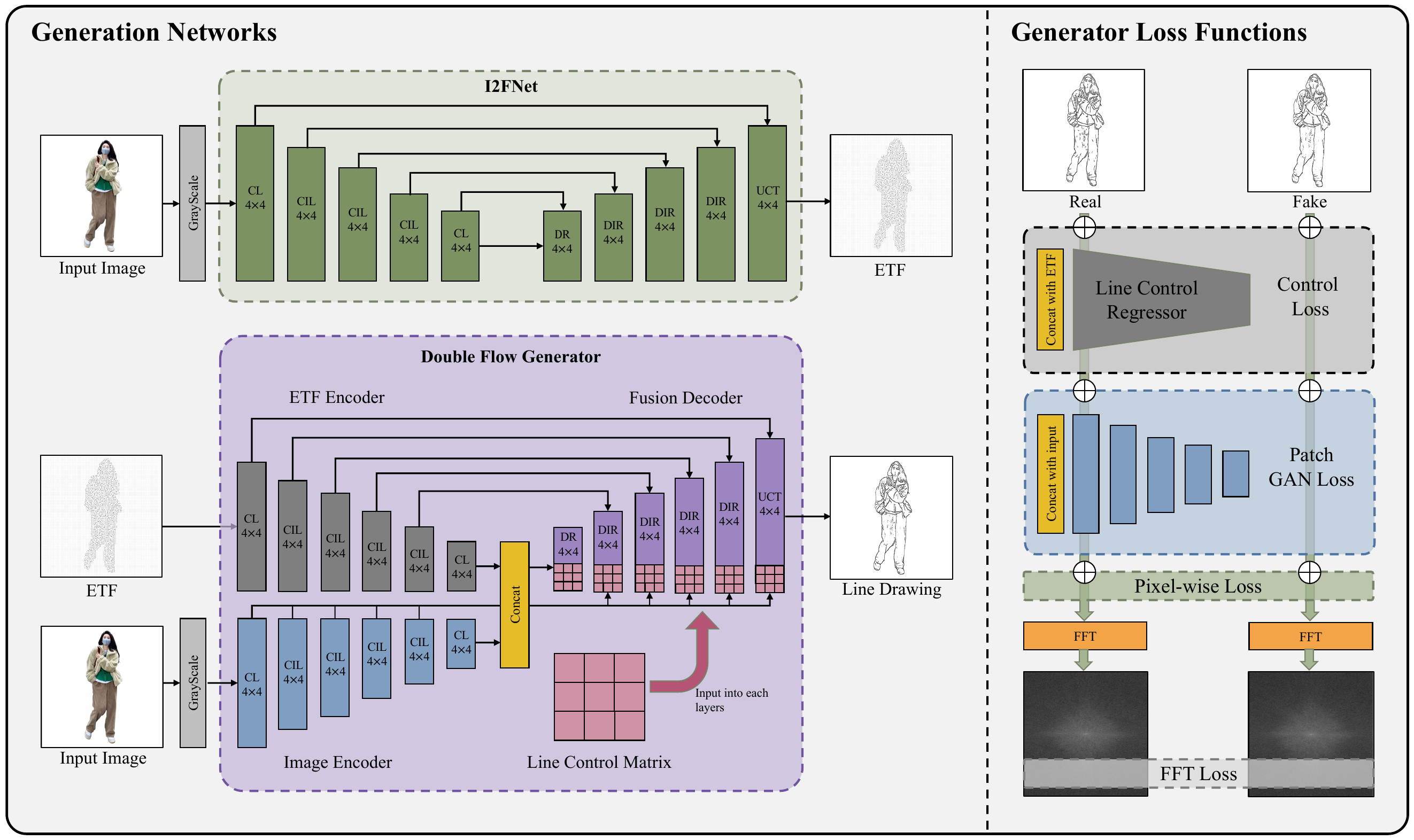}
\caption{The overall architecture of our proposed controllable artistic character line drawing generation network. \textbf{CL} denotes the operations of Convolution and LReLU. \textbf{CIL} represents the sequence of Convolution, Instance Normalization, and LReLU. \textbf{DIR} indicates Deconvolution, Instance Normalization and ReLU. \textbf{UCT} is the Upsampling, Convolution, and Tanh operations. $\bigoplus$ means taking part in the calculation of the corresponding loss.}
\label{fig_framework}
\end{figure*}

\section{Proposed Approach}
\subsection{Overview}
Since automated character line drawing generation can be formulated as a task in the domain of image-to-image translation, our goal is to perform a photo-to-line drawing transformation. That is, given an arbitrary image from source character photograph domain $\mathcal{P}$, our proposed approach outputs a corresponding image belonging to the target character line drawing domain $\mathcal{C}$ via learned mapping function $f: \mathcal{P} \rightarrow \mathcal{C}$. 
The overall architecture is schematically illustrated in Figure \ref{fig_framework}. It can be clearly seen that our model consists of an Image-to-Flow Network (Section \ref{i2fnet}) for edge flow field generation, a Double Flow Generator (Section \ref{dfg}) and a drawing discriminator (Section \ref{dd}) for edge flow guided translation from photo to line drawing, and a Control Regressor (Section \ref{cr}) for adjustment of drawing details. 

\subsection{Image-to-Flow Network (I2FNet)}
\label{i2fnet}
Actually, the direction of strokes plays an important role in determining the coherence and consistency of the lines. Kang et al. \cite{kang2007coherent} proposed the \textit{edge tangent flow (ETF)} technique providing a much better solution for finding directions of local image structure. However, the traditional method for yielding ETF is usually sensitive to user-specified parameters \cite{shi2022reference}, and requires much more expensive time cost to obtain a high-quality illustration. Therefore, it is necessary to define an efficient mapping function $f_{I2FNet}$ to perform transformation from image $\mathcal{P}$ to vector flow $\mathcal{E}$ (i.e. $f_{ETF}: \mathcal{P} \rightarrow \mathcal{E}$) in a learning based manner. 

\begin{figure}
\centering
\includegraphics[width=0.48\textwidth]{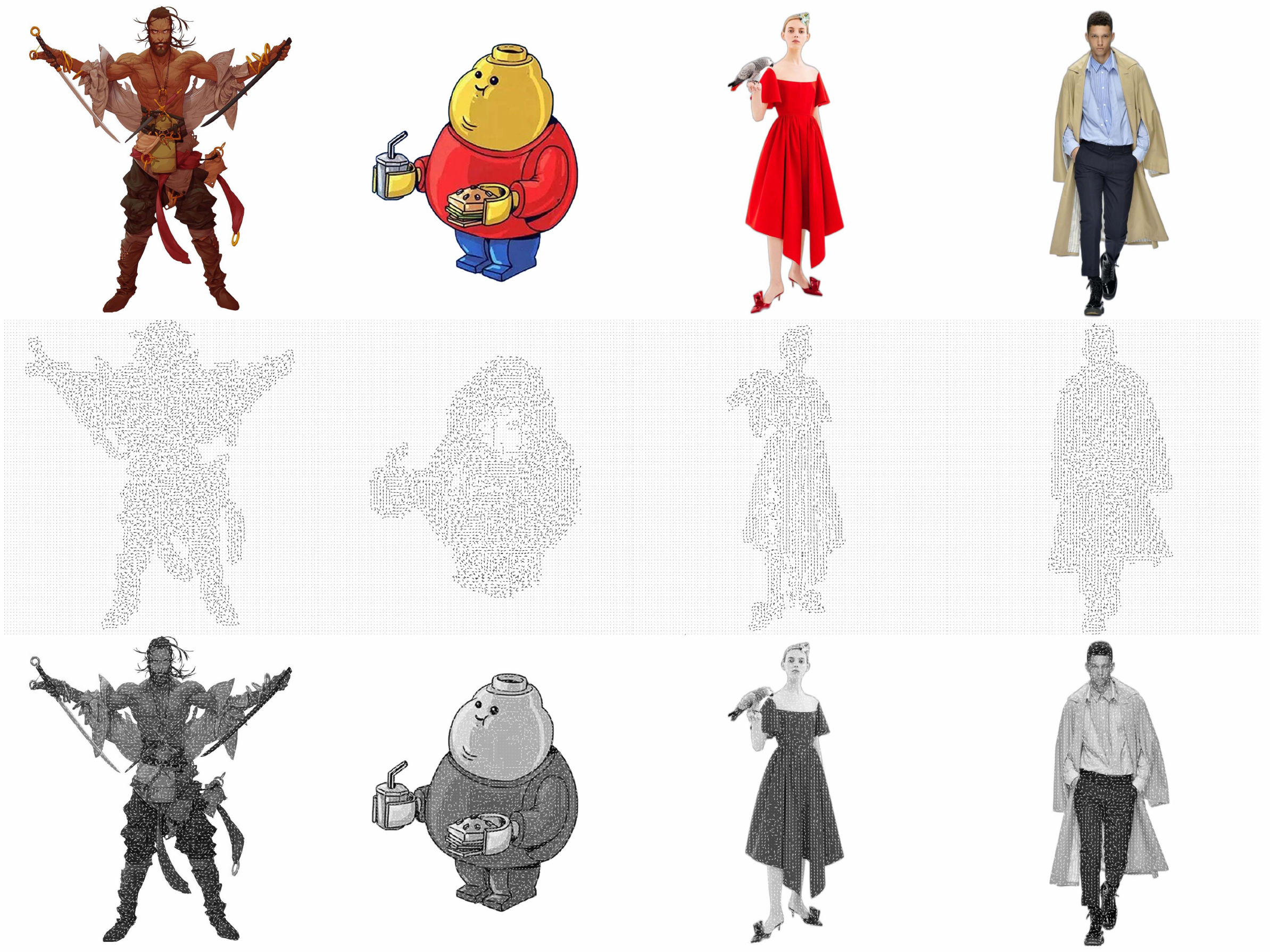}
\caption{The visualization of edge tangent field. The first row illustrates the original input images. The second row visualizes the learned edge tangent flow vector field. The images in the third row are formed by overlaying the ETF on the top of the inputs. The arrows indicate the direction of the ETF vector. }
\label{fig_eft}
\end{figure}

To this end, we propose a simple but efficient Image-to-Flow network based on the core idea of the GAN model. Specifically, the generator $G_{ETF}$ adopts a U-Net structure with five encoding layers and five decoding blocks. It begins with a transformation module that converts the input character image/photographs $p \in \mathcal{P}$ into a grayscale image, followed by hierarchical feature learning as well as four downsampling operations via CL (Convolution, LReLU) and CIL (Convolution, Instance Normalization, LReLU)  layers. Then, we use DR (Deconvolution, ReLU), DIR (Deconvolution, Instance Normalization, ReLU) and UCT (Upsampling, Convolution, Tanh) layers as encoding blocks to gradually upsample the feature maps from the last encoding layer to produce edge tangent flow vector field $G_{ETF}(p)$. For ETF discriminator $D_{ETF}$, we utilize the PatchGAN classifier \cite{isola2017image} to determine the prediction map of real/fake for each $94\times94$ patch. Figure \ref{fig_eft} displays the visualization of directions of the learned ETF vector field. 

\subsection{Double Flow Generator (DFG)}
\label{dfg}
Once the ETF field has been constructed, we can use the direction of drawings it provides to guide the generation of character lines with enhanced quality and continuity. Here, we design a Double Flow Generator network $G_{lD}$ that is fed into the original input image $p$ and the corresponding ETF vector field $e \in \mathcal{E}$ and outputs a character line drawing.      

In particular, the DFG $G_{LD}$ also adopts a U-Net framework but with two encoder branches: an ETF encoder and an image encoder. Both of them use six encoding layers to extract useful information hierarchically from input ETF and image flow. Then, at the end of the encoders, we concatenate the output feature maps and transfer the fused features into our decoder. Simultaneously, for each decoding layer, we also integrate the feature maps from corresponding encoding layers in both encoders with the help of skip connections. The final character line drawing illustration $G_{LD}(p, e)$ is created through upsampling these hierarchically-fused features.    

In order to achieve various character line drawing styles via flexible control over the line thickness and smoothness in a user-controllable manner, we introduce the following strategies. We design a Line Control Matrix(LCM) $\mathcal{A}$ which include control parameters $a$ with the same size as the input image according to the requirements, and the parameter corresponding to each part of this matrix corresponds to the style of the line corresponding to each part of the generated image. This value is constrained by us to be in the range of [0, 1] during training, where the smaller the value, the finer the corresponding generated result line, the more obvious the details, the more discontinuous and less smooth the line, while the larger the value, the coarser the corresponding generated result line, the more abstract the image, and the stronger the line smoothness and continuity. Then, we append this matrix to each decoding layer to constrain the generation of drawings to have the expected style.

\subsection{Line Drawing Discriminator }
\label{dd}
The objective of line drawing discriminator $D_{LD}$ is to distinguish the synthesized character line drawing from the ground truth counterpart. We also consider the PatchGAN structure as a discriminator to classify if each $94\times94$ is real or fake.   

\subsection{Line Control Regressor}
\label{cr}
In order to allow the control parameters to constrain the line style of the generated results as much as possible, we developed a Line Control Regressor (lCR) for the input image returning the control parameters and use this parameters to calculate loss with input control parameters. We design LCR $\mathcal{R}$ based on a Fully Convolutional Regression Network, which takes ETF, the ground truth of character line drawing and control parameters as inputs during training. 

Notably, (1) the control parameter value is in the range [0, 1]. (2) Our LCR is trained independently. The output of the trained model will be involved in the calculation of Control Loss that supervises the training of the DFG model, which is depicted in Section \ref{objective}.   

\subsection{Objective Function}
\label{objective}
In order to obtain high-quality character line drawings, we use the following four loss functions to supervise the training process. 

\textbf{Adversarial Loss.}
The DFG $G_{LD}$ attempts to generate indistinguishable character line drawing images, while the discriminator $D_{c}$ aims to differentiate the synthesized results from real ones. Therefore, we use Adversarial loss $  \mathcal{L}_{adv}^{lD}$ to encourage the output to be visually close to ground truth, which is defined as follows:

\begin{equation}
    \begin{split}
            \mathcal{L}_{adv}^{lD} &= 
            E_{p\sim \mathcal{P},c\sim     \mathcal{C}}\left [logD_{LD}(p, c) \right ] \\
            &+ E_{p\sim \mathcal{P},c\sim \mathcal{C}}\left [log(1-D_{lD}(c, G_{lD}(p, G_{ETF}(p)))) \right ] 
    \end{split}
\end{equation}

For ETF generations using I2FNet, similarly, we also take advantage of Adversarial loss to force the synthesized ETF to be similar to the target ETF domain $\mathcal{E}$.

\begin{equation}
    \begin{split}
            \mathcal{L}_{adv}^{ETF} &= 
            E_{p\sim \mathcal{P},e\sim \mathcal{E}}\left [logD_{ETF}(p, e) \right ] \\
            &+ E_{p\sim \mathcal{P}, e \sim \mathcal{E}}\left [log(1-D_{ETF}(e, G_{ETF}(p))) \right ] 
    \end{split}
\end{equation}

\textbf{Pixel-wise Loss.} In order to make the translated drawing image perfectly match the ground truth of the character line drawing from the pixel point of view, we choose to calculate the $L_{1}$ distance to measure their similarity due to the fact that (1) $L_{1}$ contributes to the generation of less blurring results \cite{yi2019apdrawinggan}\cite{isola2017image}. (2) It can stabilize the training procedure \cite{zhu2017toward}.   

\begin{equation}
   \mathcal{L}_{pixel-wise} = E_{p,e}\left [ \left \| G_{LD}(p) - e \right \|_{1}  \right ] 
\end{equation}

\textbf{Control Loss.} One of our key contributions is to perform controllable character line drawing generation via adjustment of line control parameters. To achieve this, we incorporate a Control Loss to encourage our DFG to obtain character line drawing with the expected style. The control loss $\mathcal{L}_{lc}$ can be written as,

\begin{equation}
      \mathcal{L}_{lc} = E_{p,a}\left [ \left \| \mathcal{R}(G_{LD}(p, G_{ETF}(p)), G_{ETF}(p)) - a \right \|_{1}  \right ]  
\end{equation}

\begin{figure}
\centering
\includegraphics[width=0.48\textwidth]{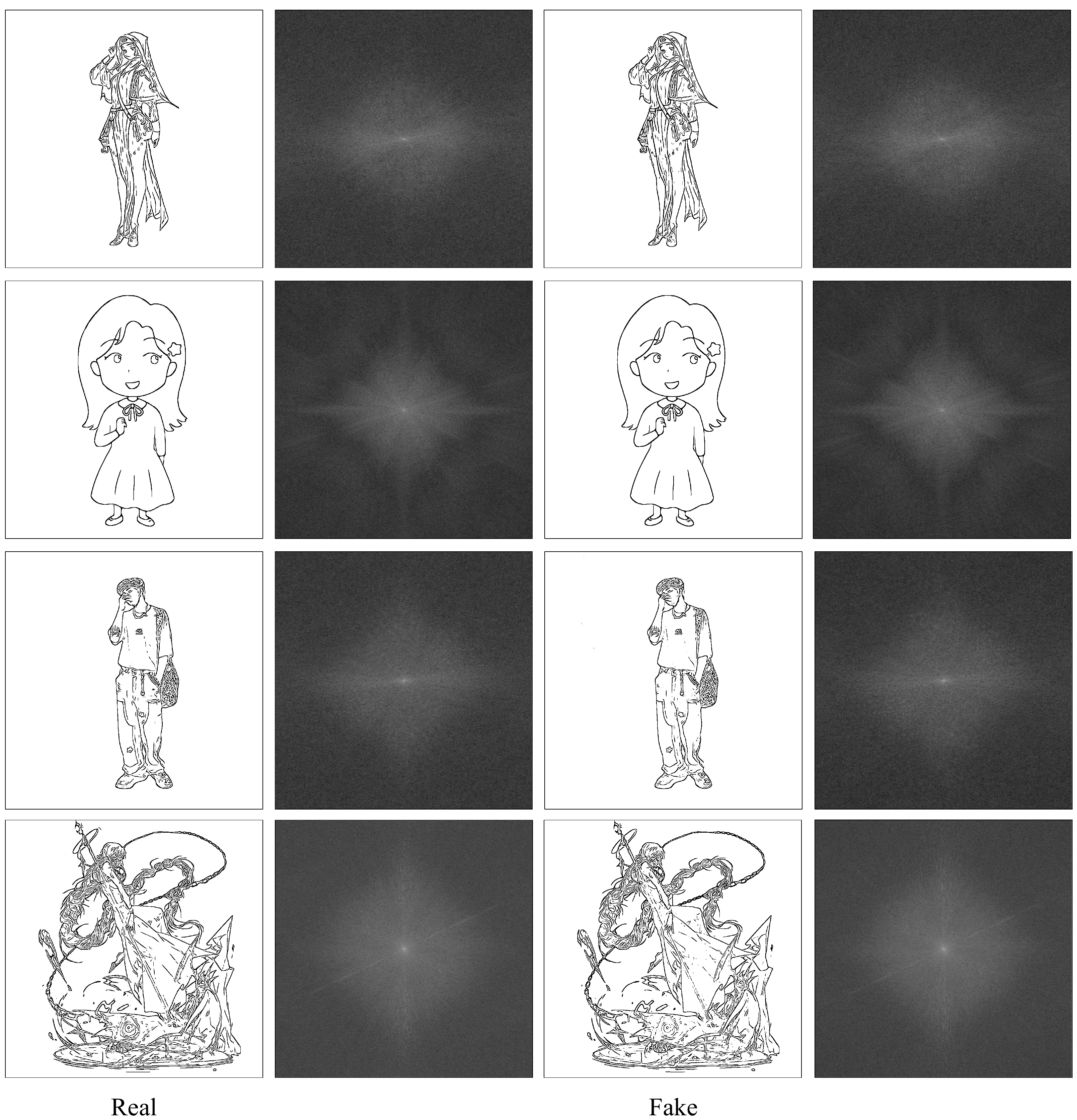}
\caption{The visualization of spectrum information for real and generated character line drawings. }
\label{fig_fft}
\end{figure}

\textbf{FFT Loss.} Theoretically, analysing character line drawing from the perspective of the frequency domain is also an ideal choice for our translation task. This is mainly because similar images should have similar spectrum features, and the edges are usually related to high frequencies. The examples visualized in Figure \ref{fig_fft} visualize the spectrum information of examples of real and synthesized character line drawings. We, therefore, design a Fast Fourier Transformation (FFT) Loss to drive the generated character line drawings to be similar to ground truth drawings in the frequency domain. The formulation of FFT loss is as follows,

\begin{equation}
    \mathcal{L}_{fft} = E_{p, c}\left [ \left \| FFT(c)-FFT(G_{LD}(p, G_{ETF}(p))) \right\|_{1} \right ]  
\end{equation}
Where $FFT(\cdot)$ represents the Fast Fourier Transformation operation.

\textbf{Total Loss.} The final objective loss function of our model is formulated as, 

\begin{equation}
    \begin{split}
    \mathcal{L}_{total} = \lambda_{adv} \mathcal{L}_{adv}^{LD} + \lambda_{pixel}\mathcal{L}_{pixel-wise} \\
    + \lambda_{lc}\mathcal{L}_{lc} + \lambda_{fft}\mathcal{L}_{fft}
    \end{split}
\end{equation}

In experiments, we set $\lambda_{adv} = 1, \lambda_{pixel} = 100, \lambda_lc = 1, \lambda_{fft} = 0.05$.

\begin{figure*}
\centering
\includegraphics[width=\textwidth]{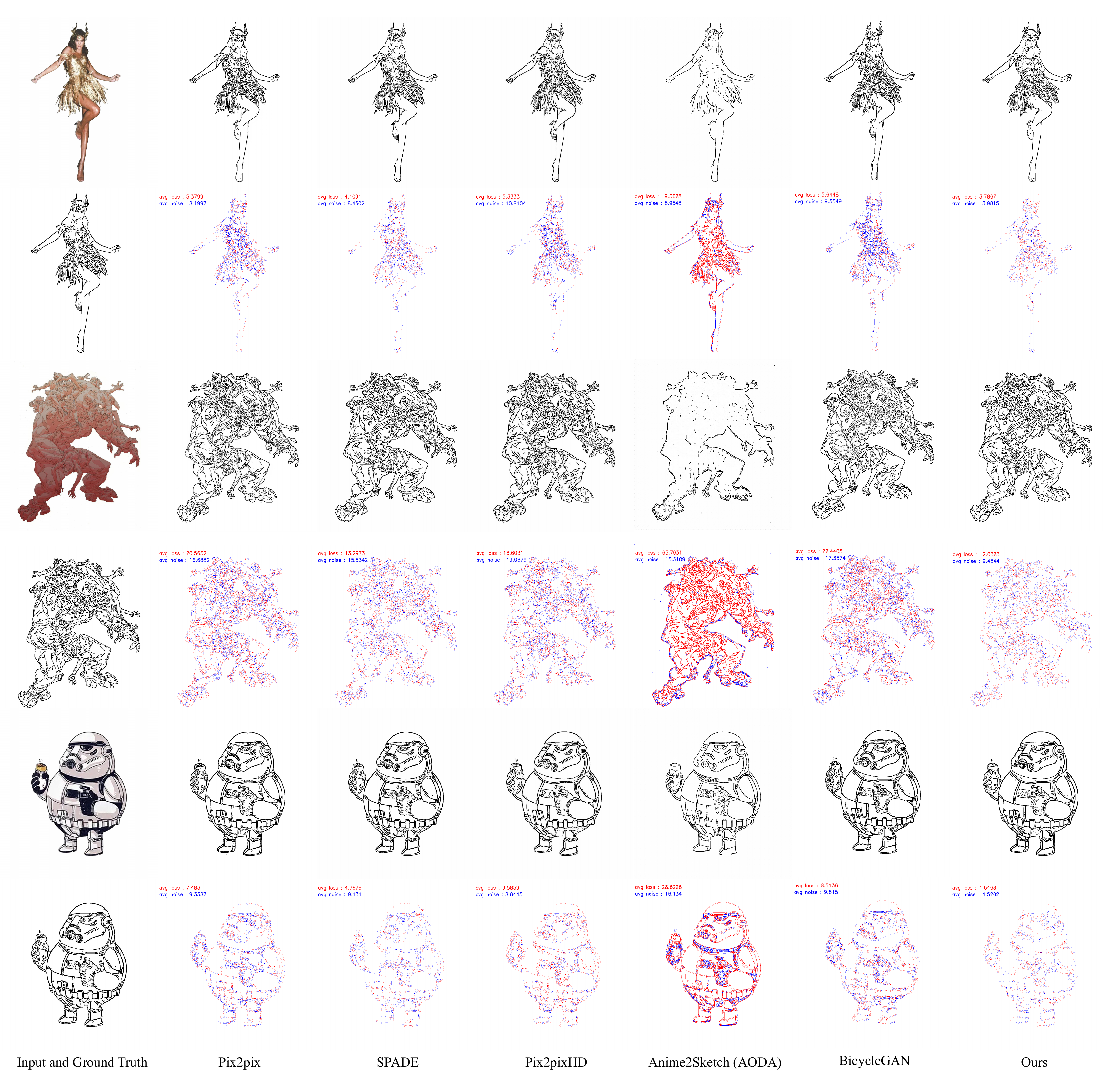}
\caption{Qualitative comparison with Pix2Pix \cite{isola2017image}, BicycleGAN \cite{zhu2017toward}, Pix2PixHD \cite{wang2018high}, SPADE \cite{park2019semantic} and Anime2Sketch \cite{xiang2022adversarial}. The odd rows are the generated character line drawings using these methods. The even rows show the difference between the generated drawings and ground truth, where the red colour represents information not learned by these methods compared to ground truth, while the blue colour denotes the generated noise. }
\label{fig_quality}
\end{figure*}

\section{Experiments}
\subsection{Implementation details}
We implement our proposed character line drawing synthesis network using Pytorch. The Adam solver with $\beta_{1}=0.5$ and $\beta_{2}=0.999$ are used to optimize our model for 200 epochs. The initial learning rate is set as 0.0002. The batch size is 2, and 1 for I2FNet. For all the methods to be compared, we choose to use the default hyperparameter settings from the original papers to train these models. All the experiments are performed on a single 24GB NVIDIA GeForce RTX 3090 GPU. And the training images are resized to $1024 \times 1024$ pixels.

\subsection{Evaluation Metrics}
To quantitatively evaluate the quality of generated character line drawings, we consider the following four metrics: (1) Fr$\acute{e}$chet Inception Distance (FID) \cite{heusel2017gans} comparing the distribution between the yielded character line drawing images and the set of grounding truth (lower value means better quality). (2) Structural Similarity Metric (SSIM) \cite{wang2004image} measuring the similarity between the generated drawing images and ground truth images (Higher scores manifest better results). (3) Peak Signal-to-noise Ratio (PSNR)  \cite{chen2020puppeteergan} evaluating the intensity difference between the prediction and ground truth (Larger values indicate smaller difference). 

\subsection{Comparison with the state-of-the-arts}
We compare our proposed approach with several state-of-the-art models, including Pix2Pix \cite{isola2017image}, BicycleGAN \cite{zhu2017toward}, Pix2PixHD \cite{wang2018high}, SPADE \cite{park2019semantic} and Anime2Sketch \cite{xiang2022adversarial}. 

For a fair comparison, we compare this only with those methods that were also developed to work with paired images and do not require additional data such as masks, labels. We use our prepared dataset and default settings to train and evaluate these networks.

\subsubsection{Quantitative evaluation}

Table \ref{tab: quantative} reports the quantitative performance evaluation of our proposed model against state-of-the-art methods. From these results, we can achieve the following observations: (1) Our character line drawing network obtains much better performance in all three measurement metrics, significantly outperforming these competing approaches. (2) The lowest FID value indicates that the distribution of our generated character line drawings is closest to that of ground truth drawings. And the highest SSIM and PSNR scores mean maximum similarity between synthesized drawings and real ones.  (3) Summarily, the quantitative evaluations verify the effectiveness of our proposed model in generating high-quality and high-fidelity character line drawings. 

\subsubsection{Qualitative comparison}
Figure \ref{fig_quality} visualizes the qualitative comparison of the results in terms of generated character line drawing itself and its difference from the ground truth. We use the red colour to represent information present in the real drawings but not in the generated ones, while the blue colour means the opposite. From these visualization results, we can conclude the following findings.

For Pix2Pix \cite{isola2017image}, BicycleGAN \cite{zhu2017toward}, Pix2PixHD \cite{wang2018high}, SPADE \cite{park2019semantic}, although they generate character line drawings with somewhat acceptable perceptual appearance, they also have many unwanted artefacts. Anime2Sketch \cite{xiang2022adversarial} yields drawings in which a great deal of detail is lost. In contrast, the results of our method are visually much closer to ground truth, compared to these models. It demonstrates that our method can not only reduce the noise/artefacts via FFT loss but also use continuous, clear and smooth lines to depict the details in the character image/photograph under the guidance of the ETF field. 

In summary, our approach performs favourably against these state-of-the-art models in terms of dealing with fine details, line quality preservation and noise reduction. 

\begin{figure*}
\centering
\includegraphics[width=\textwidth]{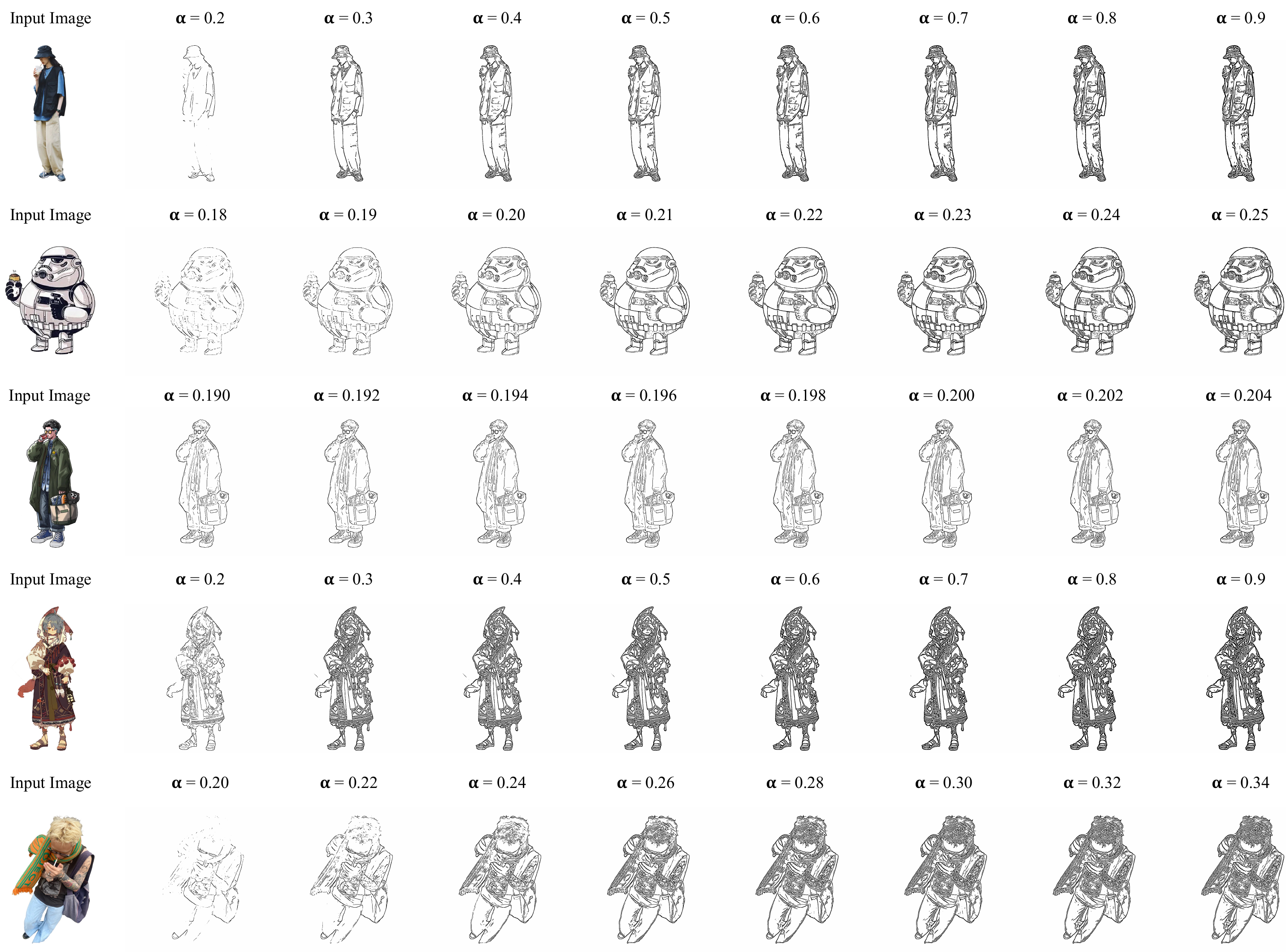}
\caption{Our model allows control over the line details, involving thickness, smoothness, continuity and stylization, of the generated character line drawings, by adjusting user-controllable parameters $\alpha$, whose value is in [0, 1].}
\label{fig_controllable}
\end{figure*}

\begin{table*}[]
\centering
\caption{Quantitative comparison of our model with several competing approaches in terms of FID, SSIM and PSNR. $\uparrow$ indicates that the higher the score, the better the performance. $\downarrow$ means that a lower value is better. \# Parames denote the number of parameters.}

\begin{tabular}{c|cccc}
\hline
 Methods  & FID $\downarrow$   & SSIM $\uparrow$  & PSNR $\uparrow$  & \#Params \\ \hline
 Pix2Pix \cite{isola2017image}   & 28.9 & 0.9370 & 19.8792 & 54.408M \\ 
 BicycleGAN \cite{zhu2017toward}  & 29.0 & 0.9308 & 19.0186 & 54.788M \\
 Pix2PixHD \cite{wang2018high}  & 27.6 & 0.9416 & 19.8944 & 182.443M \\
SPADE \cite{park2019semantic}   & 23.8 & 0.9507 & 21.1563 & 92.1M \\ 
 Anime2Sketch \cite{xiang2022adversarial}  & 58.6 & 0.8685 & 14.4924 & 54.403M \\\hline
Ours (w/o Line Control Regressor)  & \textbf{20.3} & \textbf{0.9567} & \textbf{21.8854} & 51.541M \\
 \hline
\end{tabular}
\label{tab: quantative}
\end{table*}

\subsection{User control}


The previous discussion shows that by introducing the Line Control Matrix and Line Control Regressor network, our image/photo-line drawing translation architecture can perform fine-grained control over character line drawing styles with user-specified values for control parameter $\alpha$. Figure \ref{fig_controllable} illustrates examples of character line drawings with different styles generated using different $\alpha$ for cartoon, manga and real person images/photographs. It is reasonable to draw the following conclusions: (1) By adjusting the control parameter $\alpha$, users can indeed get character line drawings with their desired styles. (2) The net effect of increasing in $\alpha$ will be a character line drawing image with an increasing amount of details. And the lines become much more clear, complete, and continuous. (3) As expected, the LCM and LCR have greatly important effects on resolving line details for character line drawing synthesis.

\begin{figure}
\centering
\includegraphics[width=0.5\textwidth]{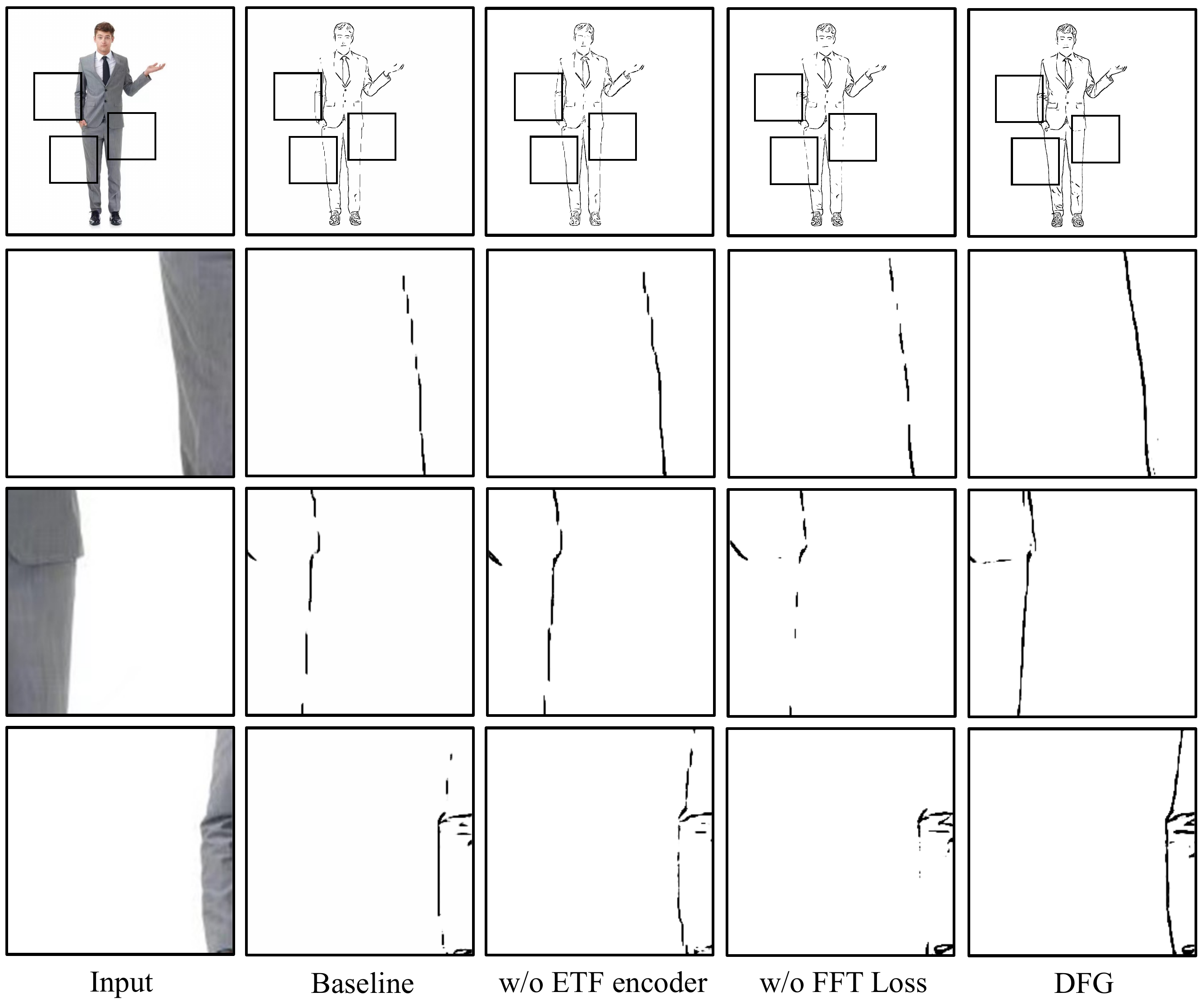}
\caption{Ablation study. Visual comparison of the effect of the core components in our translation architecture. }
\label{fig_ablation}
\end{figure}

\begin{table*}[]
\centering
\caption{Ablation study on key components in our framework.}
\begin{tabular}{c|cccc}
\hline
 Methods  & FID $\downarrow$   & SSIM $\uparrow$  & PSNR $\uparrow$   \\ \hline
 Our baseline    & 22.5 & 0.9525 & 21.2441  \\ 
 Our baseline + FFT Loss  & 22.0 & 0.9521 &  21.3189 \\
Our baseline + ETF Encoder   & 20.5 & 0.9564 & 21.7294 \\
DFG (baseline + ETF Encoder + FFT Loss)  & 20.3 & 0.9567 & 21.8854\\ 
 \hline
\end{tabular}
\label{tab: abality}
\end{table*}

\subsection{Ablation study}
We conduct ablation studies to illustrate the contribution of our proposed modules to the character line drawing synthesis. Here, we train three variants of our model: without FFT loss, without ETF encoder, and without both (our baseline). Table \ref{tab: abality} reports the quantitative performance comparison with our DFG in terms of FID, SSIM and PSNR. And Figure \ref{fig_ablation} visualizes a qualitative comparison of an example using these models. 

From these experimental results, we can achieve the following observations: (1) Removing the FFT loss and ETF encoder results in degraded performance and the reduction of similarity between generated and real character line drawings. (2) Meanwhile, we observe blurry, incoherent and incomplete lines due to the lack of guidance of stroke direction and frequency spectrum. (3) Therefore, the FFT loss and ETF encoder are essential to our character line-drawing network. They jointly encourage our method to produce visually high-quality character line drawings with clear, coherent and less noisy lines.

\section{Conclusion}
In this paper, we present a framework for controllable artistic character line-drawing generation. Three modules, namely Image-to-Flow network, Double Flow Generator and Line Controllable Regressor are well designed that contribute to synthesising higher visual-quality illustration. Experimental results demonstrate that our model can produce coherent, clear, stylistic and controllable lines. 

\bibliographystyle{IEEEtran}
\bibliography{IEEEabrv,reference}

\begin{IEEEbiography}[{\includegraphics[width=1in,height=1.25in,clip,keepaspectratio]{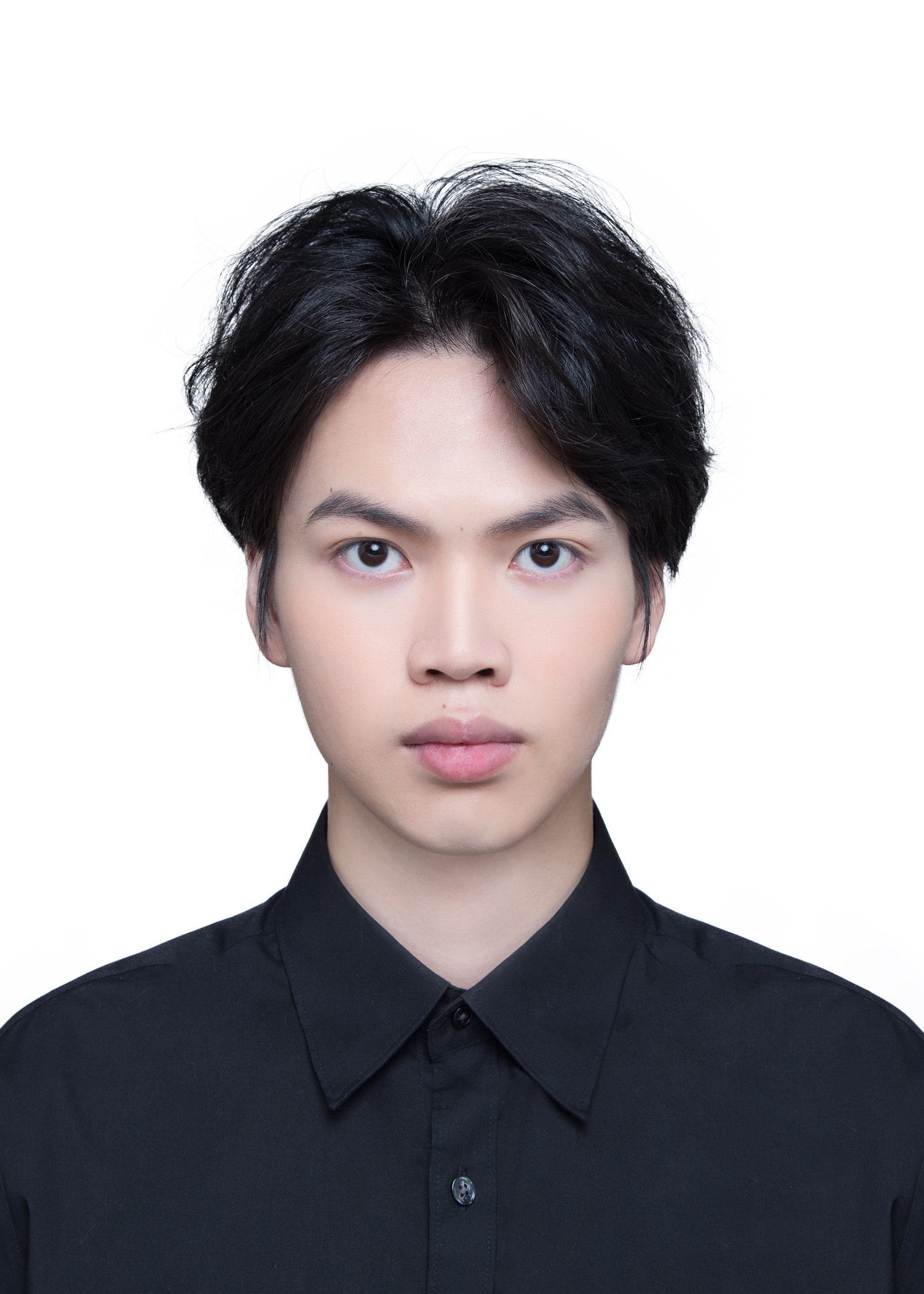}}]{Cheng-Yu Fang}

is currently pursuing the B.Eng. degree the College of Computer and Information Science at the Southwest University, Chongqing, China.  His research interests include image processing and deep learning.
\end{IEEEbiography}

\begin{IEEEbiography}[{\includegraphics[width=1in,height=1.25in,clip,keepaspectratio]{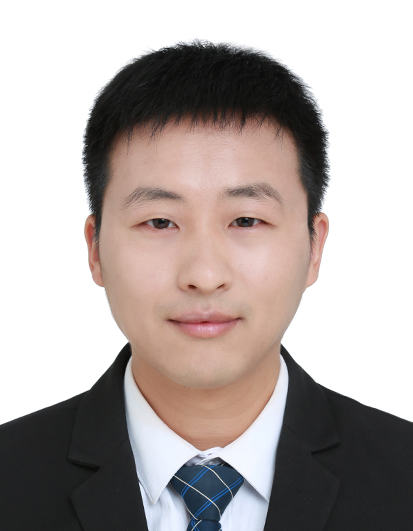}}]{Xian-Feng Han}
received the Ph.D. degree in software engineering from Tianjin University, in 2019. He is currently a lecturer with the College of Computer and Information Science, Southwest University, Chongqing, China. His research interests include 3D point cloud processing, 3D reconstruction, and machine learning.
\end{IEEEbiography}

\end{document}